\documentclass[conference]{IEEEtran}
\IEEEoverridecommandlockouts
\usepackage{cite}
\usepackage{amsmath,amssymb,amsfonts}
\usepackage{algorithmic}
\usepackage{graphicx}
\usepackage
{textcomp}
\usepackage{xcolor}
\def\BibTeX{{\rm B\kern-.05em{\sc i\kern-.025em b}\kern-.08em
    T\kern-.1667em\lower.7ex\hbox{E}\kern-.125emX}}

\usepackage{graphics}
\usepackage{url}
\usepackage{multirow}
\usepackage{siunitx}
\usepackage{bm}

\usepackage[switch]{lineno}

\usepackage{color}
\def\red#1{\textcolor{red}{#1}}
\def\blue#1{\textcolor{blue}{#1}}

\begin{document}

\title{Time-series Initialization and Conditioning for Video-agnostic Stabilization of Video Super-Resolution using Recurrent Networks
\vspace*{-1mm}
}

\author{\IEEEauthorblockN{Anonymous Authors}}
\author{\IEEEauthorblockN{Hiroshi Mori}
\IEEEauthorblockA{\textit{Toyota Technological Institute}, Nagoya, Japan \\
hiroshi.mori.ttij@gmail.com}
\and
\IEEEauthorblockN{Norimichi Ukita}
\IEEEauthorblockA{\textit{Toyota Technological Institute}, Nagoya, Japan \\
ukita@toyota-ti.ac.jp}
}

\maketitle

\vspace*{-10mm}

\begin{abstract}
A Recurrent Neural Network (RNN) for Video Super Resolution (VSR) is generally trained with randomly clipped and cropped short videos extracted from original training videos due to various challenges in learning RNNs.
However, since this RNN is optimized to super-resolve short videos, VSR of long videos is degraded due to the domain gap.
Our preliminary experiments reveal that such degradation changes depending on the video properties, such as the video length and dynamics.
To avoid this degradation, this paper proposes the training strategy of RNN for VSR that can work efficiently and stably independently of the video length and dynamics.
The proposed training strategy stabilizes VSR by training a VSR network with various RNN hidden states changed depending on the video properties.
Since computing such a variety of hidden states is time-consuming, this computational cost is reduced by reusing the hidden states for efficient training.
In addition, training stability is further improved with frame-number conditioning.
Our experimental results demonstrate that the proposed method performed better than base methods in videos with various lengths and dynamics.
\end{abstract}



\section{Introduction}
\label{section:introduction}

Video Super-Resolution (VSR) generates an SR video from 
a Low-Resolution (LR) video.
A Recurrent Neural Network (RNN) is attractive because of its advantages: high quality, high speed, and lightweight.
However, it is not easy to train RNN for a long video consisting of many frames due to various problems in learning RNN, such as vanishing and exploding gradients and the cost of computational resources.

The stability of RNN is improved if its Lipschitz constant is less than 1~\cite{DBLP:conf/iclr/MillerH19,DBLP:conf/icml/MhammediHRB17,DBLP:conf/icml/VorontsovTKP17,DBLP:conf/icml/JoseCF18}.
As with these general RNNs, RNNs for videos can also be stabilized by the Lipschitz constraint (e.g., video denoising~\cite{DBLP:journals/pami/TanaySMDTLS23} and VSR~\cite{DBLP:conf/cvpr/ChicheWFS22}).
However, these methods have difficulty in VSR for various video properties such as video length and dynamics.
Furthermore, 
the cost of computational resources is still a big issue in training videos because the memory of videos is much larger than that of other media, such as texts.

\begin{figure}[t]
\vspace*{-7mm}
\begin{center}
\includegraphics[width=\columnwidth]{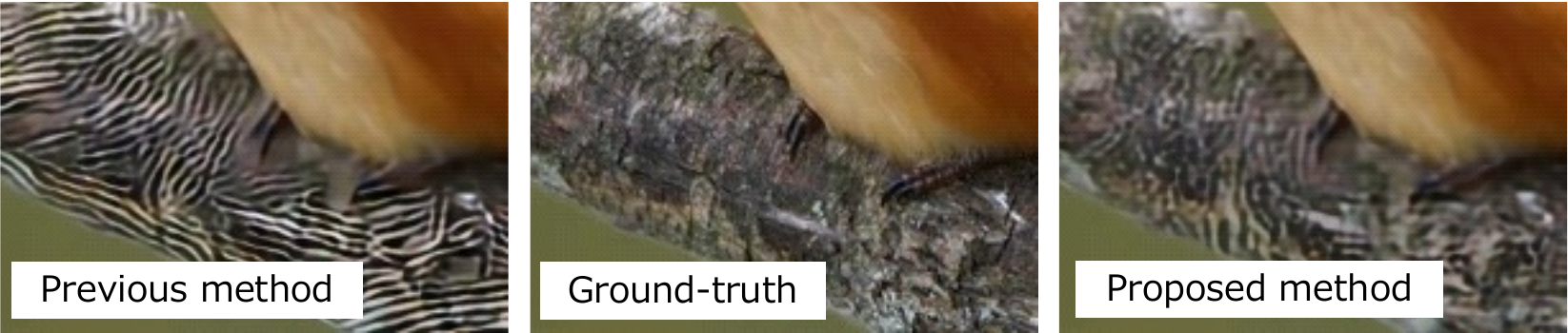}\\
(a) Static video\\
\includegraphics[width=\columnwidth]{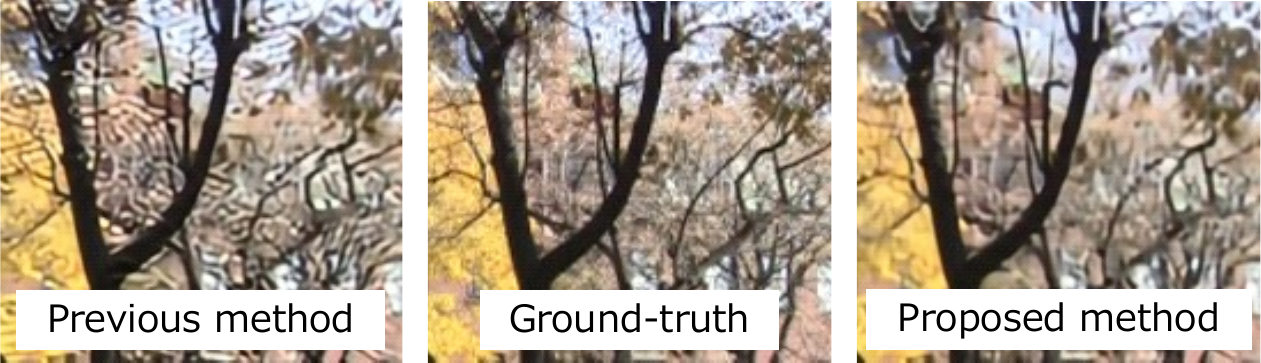}\\
(b) Dynamic video
\end{center}
\vspace*{-2mm}
\caption{Difference between previous~\cite{DBLP:conf/cvpr/SajjadiVB18,DBLP:conf/cvpr/ChanWYDL21} and our VSR methods.
}
\label{fig:top}
\vspace*{-2mm}
\end{figure}

This memory problem is generally avoided by training RNN with short small videos that are temporally clipped and spatially cropped from long videos.
However, this training scheme
causes a domain gap between RNN's hidden states computed in short and long videos.
This domain gap makes it difficult for RNN trained with short videos to super-resolve long videos, as validated in~\cite{DBLP:conf/cvpr/ChicheWFS22,DBLP:conf/cvpr/ChanZXL22}.

The following contributions of this paper improve the stability of RNN training by reducing a domain gap between video clips between the training and inference stages:
\begin{itemize}
    \item Our preliminary experiments confirm that video degradation in VSR using RNN (e.g., leftmost examples in Fig.~\ref{fig:top}) is changed depending on several video properties.
    \item For video-agnostic VSR, our method stabilizes VSR (e.g., rightmost examples in Fig.~\ref{fig:top}) by training the VSR network with various RNN hidden states changed depending on the video properties.
    \item Even with such a variety of RNN hidden states, a VSR network is efficiently trained by reusing the hidden states among multiple training video clips.
    \item Training stability is further improved with frame-number conditioning.
\end{itemize}

\section{Related Work}
\label{section:related}

\subsection{VSR Networks~\cite{DBLP:conf/cvpr/NahTGBHMSL19,DBLP:conf/eccv/FuoliHGTREKXLXW20}}
\label{subsection:vsr}

\noindent
{\bf Sliding window-based VSR~\cite{DBLP:conf/cvpr/WangCYDL19,DBLP:conf/cvpr/TianZ0X20,DBLP:journals/ijcv/XueCWWF19,DBLP:conf/cvpr/HarisSU19,DBLP:conf/cvpr/HarisSU20}:}
$t$-th LR frame (denoted by $L_{t}$) is super-resolved by its neighbor frames
(e.g., $\{ L_{t-T} \cdots, L_{t+1}, \cdots L_{t+T} \}$).
These neighbor frames are temporally slid to super-resolve the next frame.
This sliding window-based approach has the following problems:
(i) quality degradation due to a limited number of temporal receptive fields~\cite{DBLP:conf/cvpr/ChanWYDL21}, (ii) computational speed reduction due to redundant processing of the same neighbor frames for super-resolving different frames~\cite{DBLP:conf/cvpr/SajjadiVB18}, and (iii) flickering artifacts due to independent processing between continuous frames~\cite{DBLP:conf/cvpr/SajjadiVB18,DBLP:conf/cvpr/ChanZXL22a,DBLP:conf/iccv/YiWJJ019}.


\noindent
{\bf RNN-based VSR~\cite{DBLP:conf/cvpr/SajjadiVB18,DBLP:conf/cvpr/ChanWYDL21,DBLP:conf/cvpr/ChicheWFS22,DBLP:conf/cvpr/ChanZXL22a,DBLP:conf/iccvw/FuoliGT19,DBLP:conf/eccv/IsobeJGLWT20}:}
RNNs, including unidirectional RNNs~\cite{DBLP:conf/cvpr/SajjadiVB18,DBLP:conf/cvpr/ChicheWFS22,DBLP:conf/iccvw/FuoliGT19,DBLP:conf/eccv/IsobeJGLWT20} and bidirectional RNNs~\cite{DBLP:conf/cvpr/ChanWYDL21,DBLP:conf/cvpr/ChanZXL22a}, are widely applied to VSR networks.
In FRVSR~\cite{DBLP:conf/cvpr/SajjadiVB18}, which is a unidirectional RNN, an SR frame at $t-1$ (denoted by $S_{t-1}$) is warped by an upscaled flow image between $t-1$ and $t$ to use this warped SR image for super-resolving $L_{t}$.
BasicVSR~\cite{DBLP:conf/cvpr/ChanWYDL21}, which is a bidirectional RNN, is designed with a good combination of existing modules based on the detailed empirical verification,

Unlike the sliding window-based approach, in RNN, (i) a temporal receptive field is not limited, (ii) since each frame is fed into RNN only once, redundant processing of the same frames is avoided, and (iii) by recurrently updating and sharing the hidden state through frames, temporal flickering artifacts can be suppressed.
Since all of these advantages of RNN are beneficial for real applications, this paper focuses on improving VSR using RNN.
While our experiments are conducted with FRVSR~\cite{DBLP:conf/cvpr/SajjadiVB18} and BasicVSR~\cite{DBLP:conf/cvpr/ChanWYDL21}, which are selected as typical unidirectional and bidirectional RNN-based methods, our method is applicable to any RNN-based VSR methods.


\noindent
{\bf Transformer-based VSR~\cite{DBLP:journals/corr/abs-2201-12288,DBLP:journals/corr/abs-2106-06847}:}
Transformer~\cite{DBLP:conf/nips/VaswaniSPUJGKP17} is also applied to video processing~\cite{DBLP:journals/corr/abs-2201-12288,DBLP:journals/corr/abs-2106-06847,DBLP:conf/nips/ShiGXWYD22,DBLP:conf/nips/LiangFXRIGC0TG22,DBLP:conf/eccv/ZhangFL22,DBLP:conf/icml/LinCHWYZDZTG22} such as VSR.
For example, in VSR-Transformer~\cite{DBLP:journals/corr/abs-2106-06847}, a spatial-temporal convolutional attention layer exploits the locality and spatiotemporal relationships through different layers.
In addition,
a bidirectional optical flow-based feed-forward layer improves feature propagation and alignment between frames.
While efficiency can be improved as shown in~\cite{DBLP:conf/nips/LiangFXRIGC0TG22}, Transformer still has difficulty in efficient VSR.

\subsection{Training Stability in RNN-based VSR}
\label{subsection:stable_vsr}

RNNs
can be stabilized by the Lipschitz constraint.
In~\cite{DBLP:conf/iclr/MillerH19}, for example, the Lipschitz constraint is satisfied by computing the SVD of the weight matrix $W$ and thresholding its singular values are less than 1.

While enforcing the Lipschitz constraint in CNNs is not easy, several approaches are proposed for
non-recurrent CNNs, mainly by analyzing the singular values of a matrix reshaped from convolutional kernels~\cite{DBLP:conf/iclr/SedghiGL19,DBLP:conf/iclr/MiyatoKKY18,DBLP:conf/iclr/SanyalTD20,DBLP:journals/ml/GoukFPC21}.

With the aforementioned success of stabilizing the training of RNNs and CNNs, RNN-based VSR with convolutional layers is also explored.
In video denoising~\cite{DBLP:journals/pieee/Werbos90}, typical artifacts caused by video
RNNs are suppressed by the Stable Rank Normalization of the Layer extended from SRN~\cite{DBLP:conf/iclr/SanyalTD20}.
In~\cite{DBLP:conf/cvpr/ChicheWFS22}, the difficulty in the straightforward use of the Lipschitz constraint in an RNN-based VSR network is empirically verified; the soft Lipschitz constraint cannot suppress artifacts, while the Lipschitz constraint on all conv layers is better in terms of artifact suppression but reduces the VSR ability.
This problem is resolved by enforcing the Lipschitz constraint only on middle conv layers, even if this network is not globally Lipschitz constrained.
While this method~\cite{DBLP:conf/cvpr/ChicheWFS22} is good at VSR of long static videos, videos with different video properties cannot be super-resolved well.

\section{Preliminary Experiments}
\label{section:preliminary}

This section shows experimental results that reveal changes in the VSR stability depending on video properties related to video length, dynamics, and complexity.
While all experiments shown in Sec.~\ref{section:preliminary} were conducted with FRVSR and BasicVSR, only the results of FRVSR are shown in Sec.~\ref{section:preliminary} because 
both have similar results.

\begin{figure}[t]
\begin{center}
\includegraphics[width=\columnwidth]{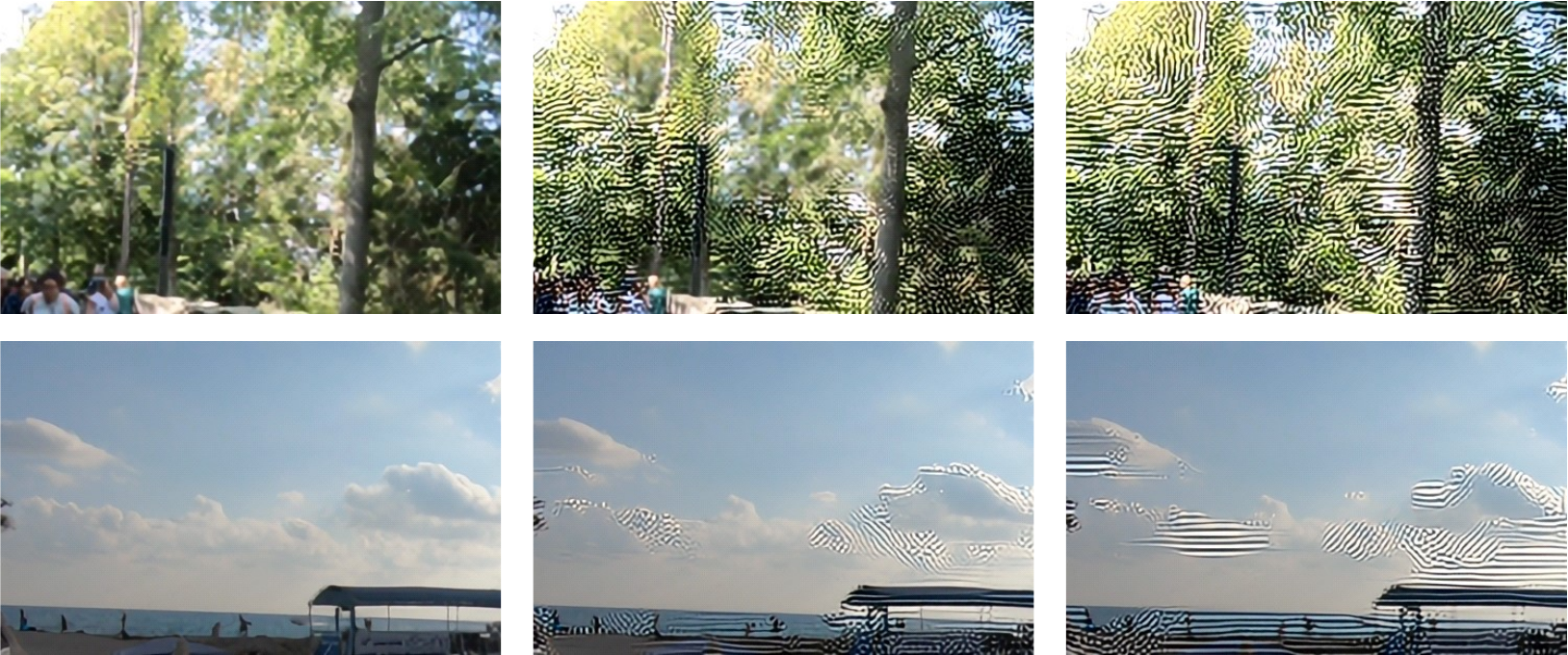}\\
$t=5$ frames
\hspace*{8mm}
$t=100$ frames
\hspace*{6mm}
$t=300$ frames
\end{center}
\vspace*{-3mm}
\caption{
The effects of the video length and texture density.
$t$ denotes the frame number in which images in the same column are reconstructed.
}
\label{fig:p_ex_1}
\end{figure}

\noindent
{\bf Video length and texture density:}
LR frames with dense and less textures were manually selected.
Each selected frame is copied and temporally concatenated to generate an LR video to fix each video's properties.

In comparison between VSR videos with dense and less textures (Fig.~\ref{fig:p_ex_1} ), many artifacts are produced in the former.
Such artifacts are observed around textures (e.g., in clouds), even in the latter.
Regarding the video length, artifacts are increased as the number of frames is greater.

\begin{figure}[t]
\begin{center}
\includegraphics[width=\columnwidth]{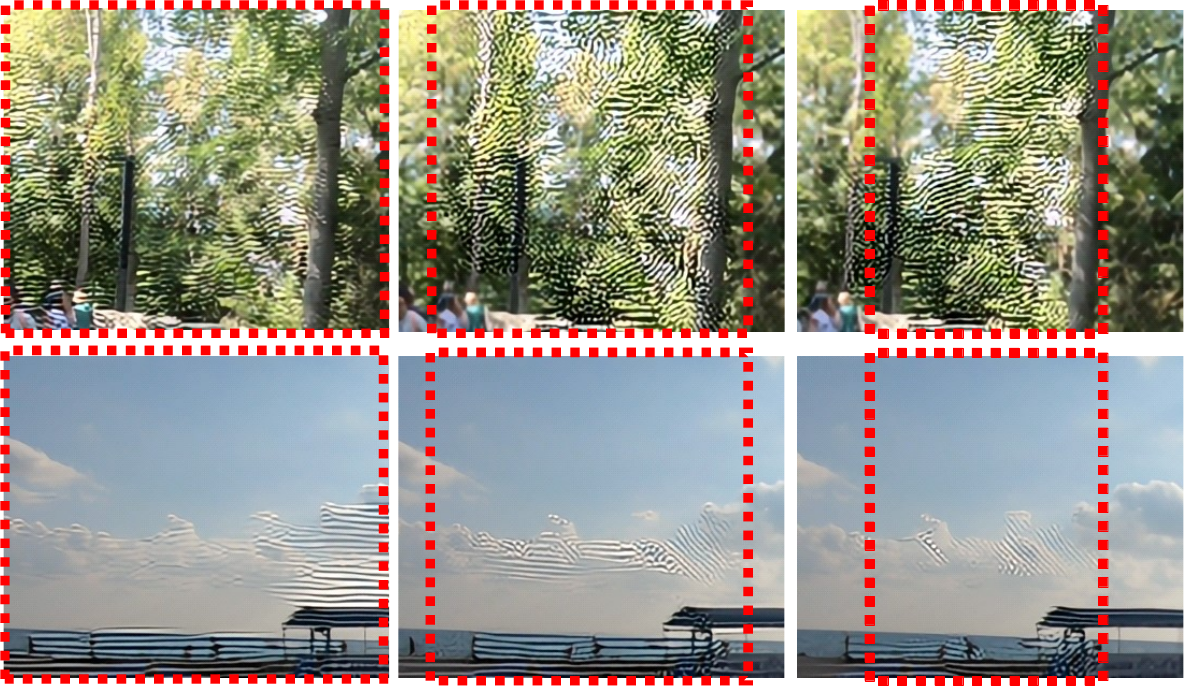}\\
$s=1$ pixel
\hspace*{8mm}
$s=4$ pixels
\hspace*{8mm}
$s=16$ pixels
\end{center}
\vspace*{-3mm}
\caption{
The effects of the motion magnitude and object disappearances.
$s$ denotes the pixel number of the window sliding.
Each red rectangle indicates the overlap between the sliding windows (i.e., pixels observed in all the sliding windows).
}
\label{fig:p_ex_2}
\end{figure}

\noindent
{\bf Motion magnitude and object disappearances:}
Given a LR frame with $w \times h$ pixels, a window with $w' \times h'$ pixels, where $w > w'$ and $h > h'$, is cropped and slid within the original frame to generate a camera motion artificially in a generated video consisting of the cropped windows.
By changing the sliding size, the motion magnitude is correctly determined in each video.
In each video, some objects appear and disappear due to the sliding window scheme.

The VSR results are shown in Fig.~\ref{fig:p_ex_2}.
We can see that artifacts are decreased as the motion magnitude becomes larger.
It can also be seen that such artifacts are not observed in objects that disappear and appear in the sliding windows.

\begin{figure}[t]
\begin{center}
\includegraphics[width=0.9\columnwidth]{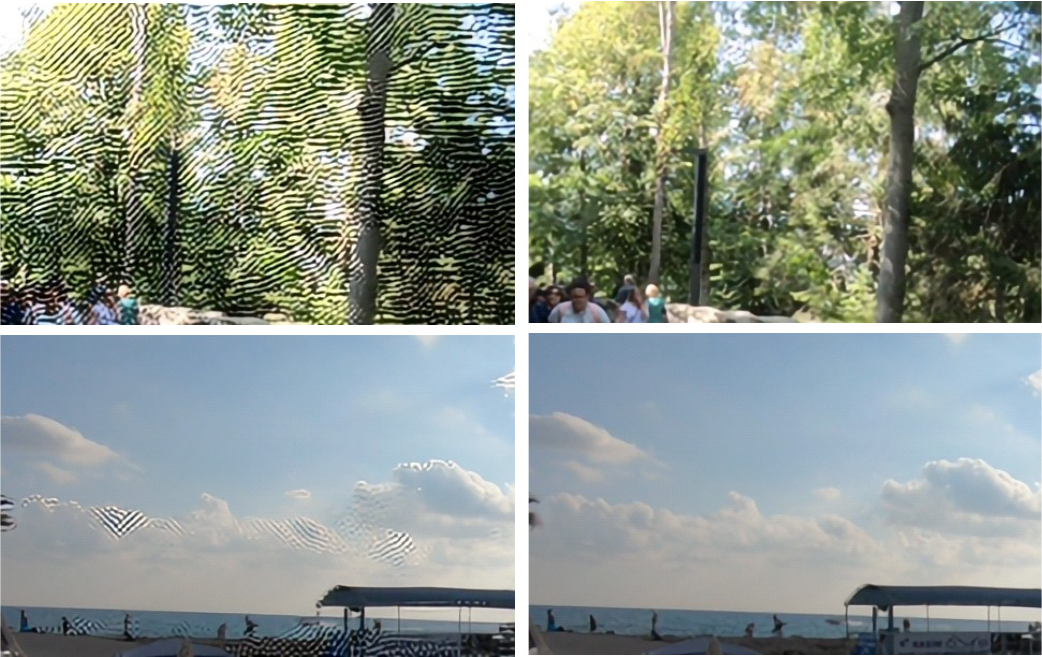}\\
Small intensity change
\hspace*{5mm}
Large intensity change
\end{center}
\vspace*{-3mm}
\caption{Preliminary experimental results. The effects of the intensity change.
All these images are $t = 300$ frames.
}
\label{fig:p_ex_4}
\end{figure}

\noindent
{\bf Intensity change:}
LR frames are randomly selected from 
all LR videos.
Each selected frame is copied and temporally concatenated to generate each LR video so that all the video properties are fixed in each video, as with the aforementioned experiments.
Continuous frames in the generated video are gradually darkened and brightened using gamma correction.
The degree of the intensity change is determined with the max and min values of $\gamma$.

While most pixels are reconstructed well, several artifacts are produced in the densely-textured image with a small intensity change, as shown in the red rectangle.
In comparison with the results 
shown in Fig.~\ref{fig:p_ex_1}, the amount of artifacts is clearly reduced by a large intensity change.

\noindent
{\bf Summary:}
Through all the experiments, artifacts are suppressed in regions where temporally correct pixel correspondences are difficult to find;
for example, long frames and less textures in Fig.~\ref{fig:p_ex_1}, large motions and object disappearances in Fig.~\ref{fig:p_ex_2}, and large intensity changes in Fig.~\ref{fig:p_ex_4}.
We interpret this result as follows: If pixel correspondence estimation fails in a pixel of interest,
hidden states representing this pixel are initialized in RNN.
This initialization resolves accumulated errors in the hidden states even though useful information is also accumulated in the hidden states.

Based on the aforementioned experimental results, we focus on how to improve the hidden states to reduce the artifacts.
The straightforward way is to initialize the hidden states when or before the error is accumulated in the hidden states.
However, it is difficult to find whether the error is accumulated.
Furthermore, the irrationalistic initialization of the hidden states decreases the effect of RNN.

For maintaining the hidden states with less error, this paper proposes the 
video-agnostic
RNN training strategy in which a variety of hidden states are trained to reduce error accumulation in the hidden states in inference due to a domain gap between the training and inference stages.

\section{Backpropagation-Through-Time and its Problems for Video Processing}
\label{section:BPTT}

\begin{figure}[t]
\begin{center}
\includegraphics[width=\columnwidth]{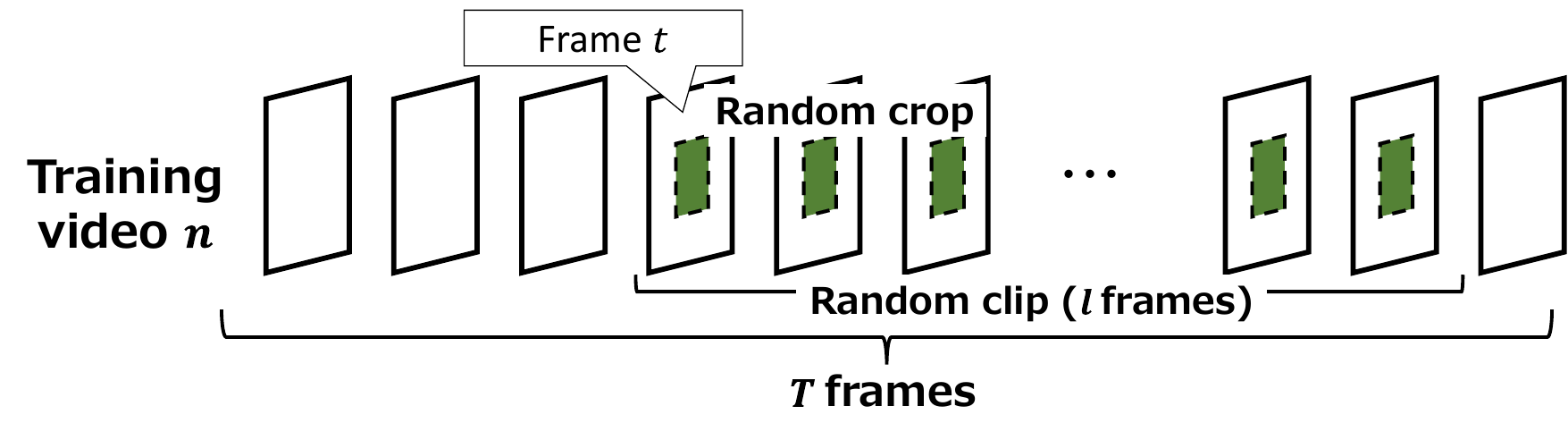}\\
Step 1: Temporal random clip and spatial random crop.\\
\vspace*{2mm}
\includegraphics[width=\columnwidth]{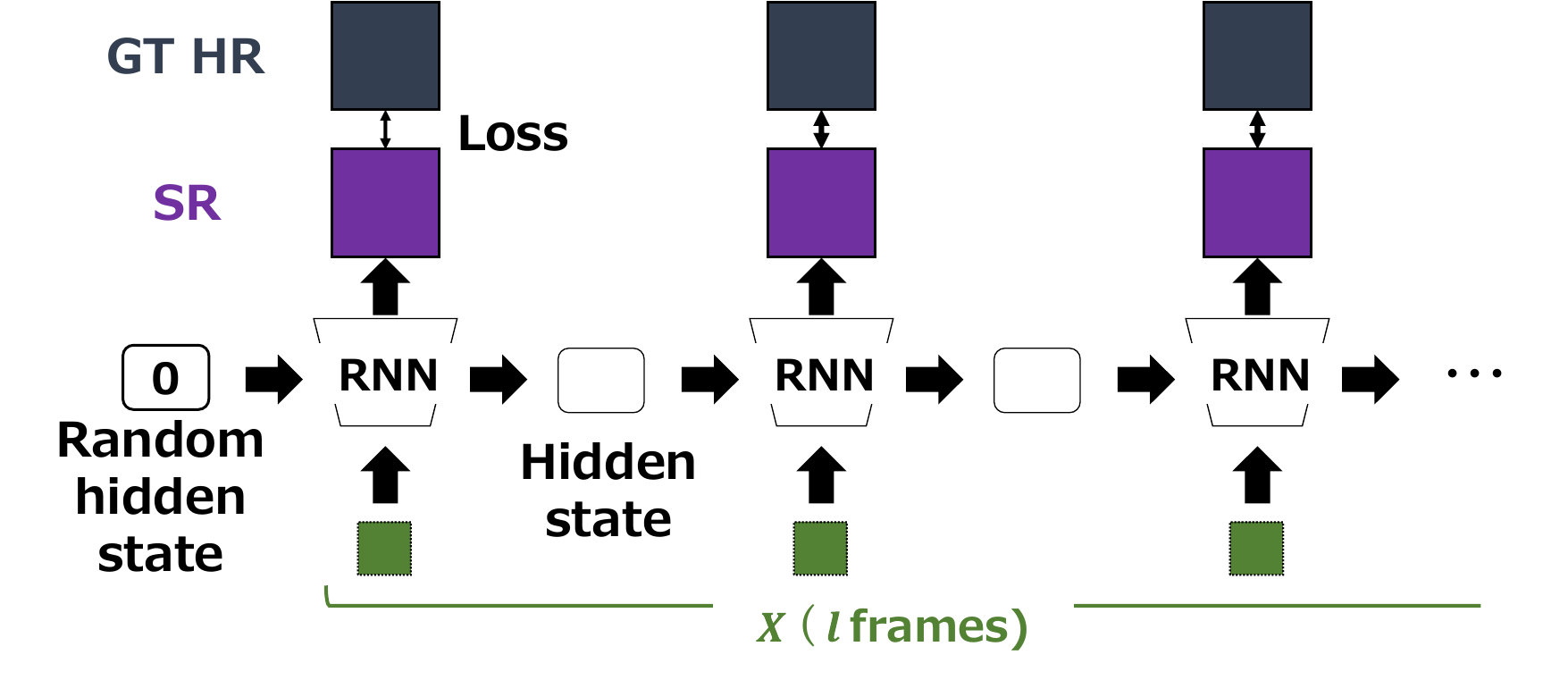}\\
Step 2: RNN training with a random initial hidden state.
\end{center}
\vspace*{-2mm}
\caption{Random-Initialization BPTT
for videos.
}
\label{fig:RIBPTT}
\end{figure}

In Backpropagation-Through-Time (BPTT) for training RNN, RNN is unrolled over an input sequential data.
BPTT for long sequential data has two difficulties: instability in training (i.e., vanishing and exploding gradients) and the computational cost (i.e., huge memory usage).

These difficulties can be avoided by truncated BPTT~\cite{DBLP:conf/uai/AicherFF19,DBLP:conf/slt/TangG18,DBLP:journals/corr/TallecO17a,DBLP:journals/neco/WilliamsP90}.
In general truncated BPTT, given each training temporal data,
the sets of hidden states
are obtained by one-time feed-forward processing of RNN and stored once.
Then, short clips are temporally clipped from the training data, and BPTT is applied to each short clip with the stored hidden states.
Since the number of recurrent steps is smaller in the short clip than in the original long data, BPTT with each short clip (i.e., truncated BPTT) can be performed with a lower memory cost.
During this training scheme, redundant feed-forward processes for computing the outputs and hidden states are avoided by using the stored outputs and hidden states.

In VSR, in addition to the aforementioned temporal clipping, a small window is cropped from each frame in a video (i) to reduce the memory cost
and (ii) to improve model generalizability based on data augmentation with small windows cropped from different locations.

\noindent
{\bf Problem of memory efficiency:}
However, it is difficult to apply truncated BPTT to video processing networks.
Since the memory of videos is much larger than that of other media, such as texts, it is impossible to store the hidden states of all frames in all videos (in a GPU memory) for truncated BPTT.
Furthermore, the memory usage required for video processing increases due to random window cropping in the frames.
For maximizing the effect of data augmentation using many overlapping cropped windows, the memory usage is much larger than that for the original frames.

\noindent
{\bf Problem of accuracy:}
Instead of storing the hidden states,
a random hidden state can be used for initialization.
In the training scheme of prior RNN-based VSR methods~\cite{DBLP:conf/cvpr/SajjadiVB18,DBLP:conf/cvpr/ChanWYDL21}, after each short clip is produced by temporal random clipping and spatial random cropping (Step 1 of Fig.~\ref{fig:RIBPTT}),
the hidden states are just randomly initialized, as shown in Step 2 of Fig.~\ref{fig:RIBPTT}.
This training strategy is called truncated Random-Initialization BPTT (truncated RI-BPTT, in short) in this paper.
However, a domain gap between ``this randomly initialized hidden state used in {\em training short videos}'' and ``a sequentially accumulated hidden state in {\em inference for a long video}'' degrades the VSR quality in inference.

\section{Proposed Method}
\label{section:proposed}

\subsection{Truncated Partial-Initialization BPTT}
\label{subsection:PIBPTT}

\begin{figure}[t]
\begin{center}
\includegraphics[width=\columnwidth]{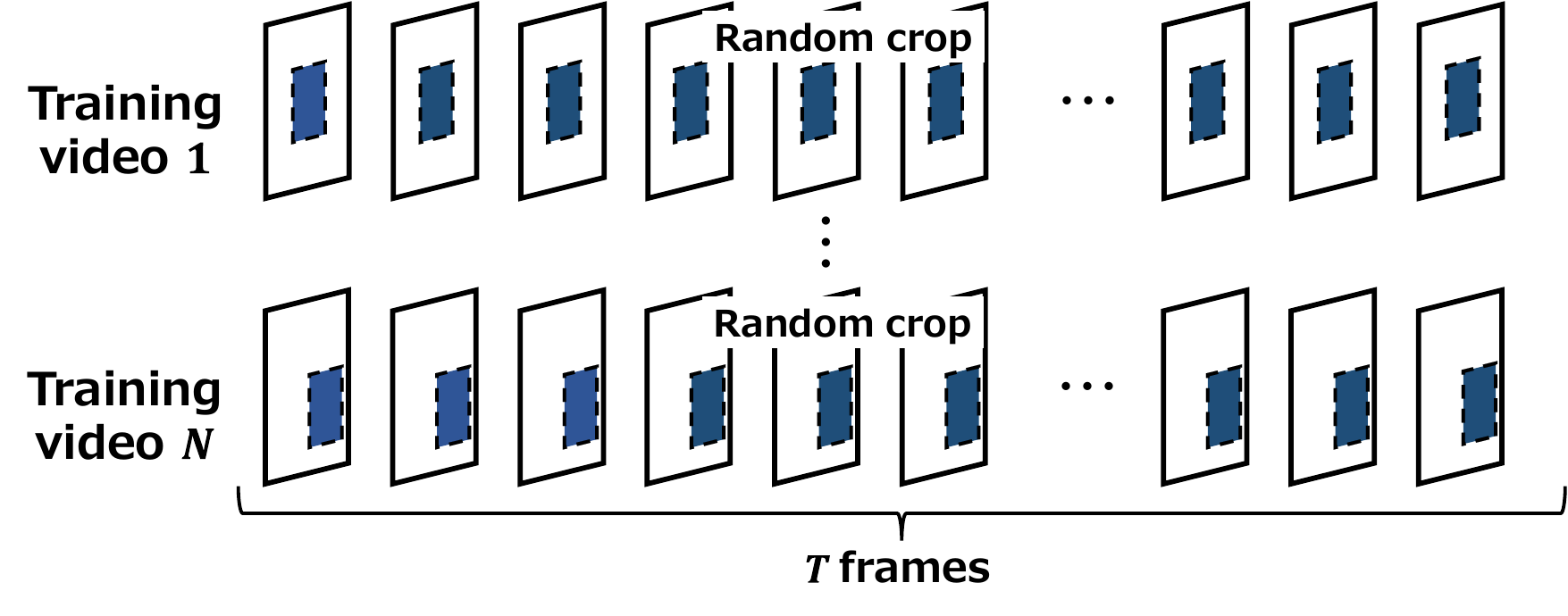}\\
\vspace*{-1mm}
\flushleft{Step 1: Spatial random crop for all $N$ training videos.}\\
\vspace*{4mm}
\includegraphics[width=\columnwidth]{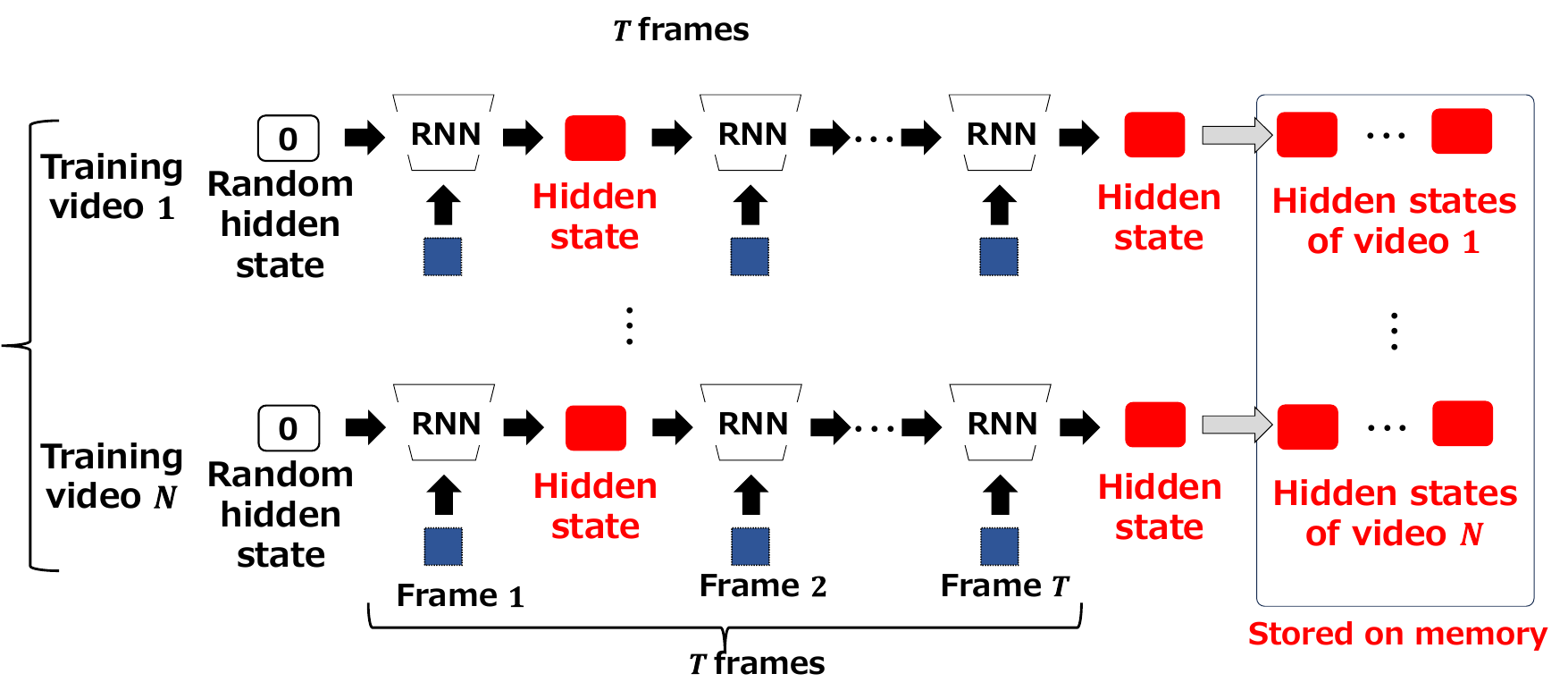}\\
\vspace*{-1mm}
\flushleft{Step 2: Hidden state computation and storage for all $N$ training videos.}\\
\vspace*{4mm}
\includegraphics[width=\columnwidth]{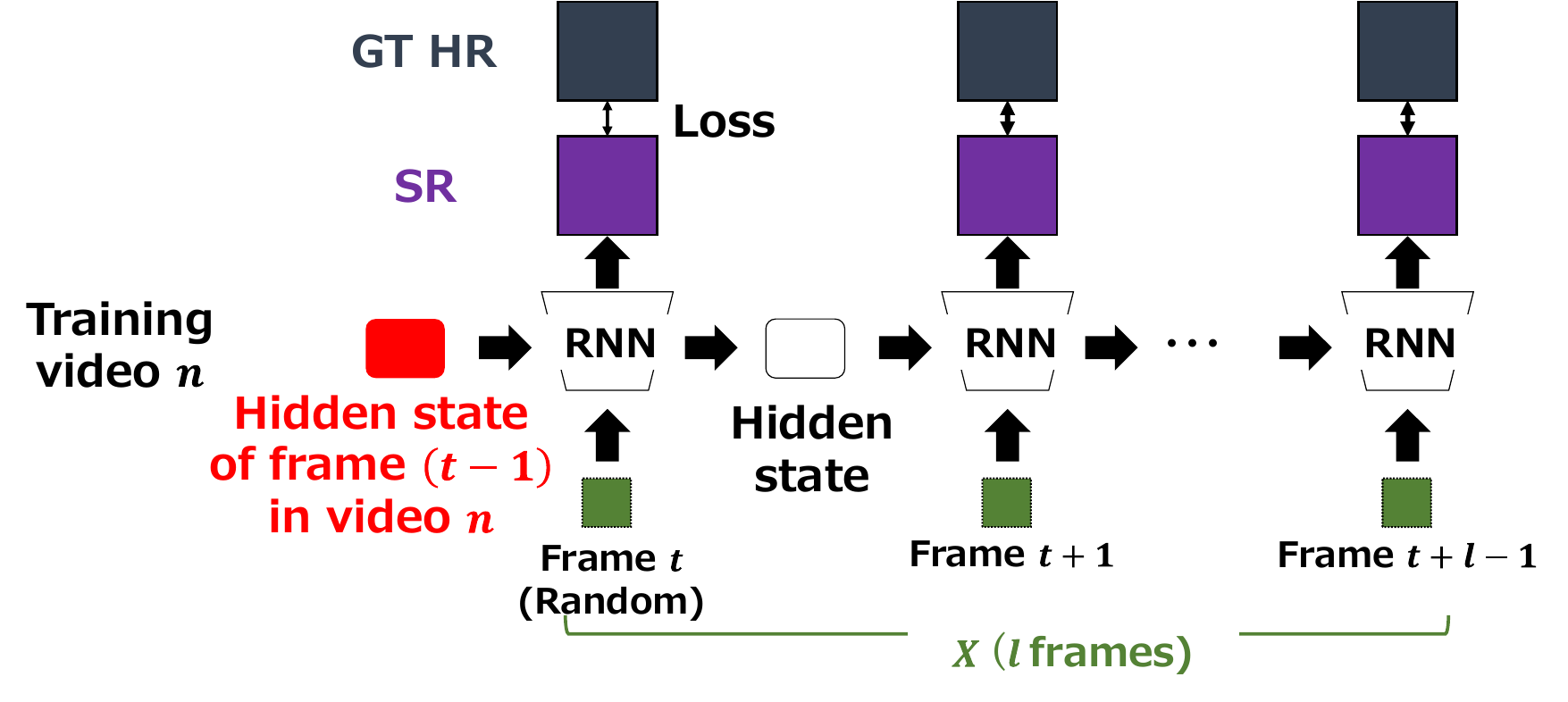}\\
\vspace*{-1mm}
\flushleft{Step 3: Truncated RNN with the hidden state given by Step 2.}
\end{center}
\vspace*{-2mm}
\caption{Partial-Initialization Backpropagation-Through-Time (PI-BPTT) for videos. Step 3 is repeated $NR$ times in each epoch. $N$ denotes the number of the training videos. Random temporal clipping for truncated RNN is repeated $R$ times in each video.
}
\label{fig:PIBPTT}
\end{figure}

To resolve the problems described at the end of Sec.~\ref{section:BPTT}, our proposed method is designed for a good trade-off between memory efficiency and accuracy.
In terms of memory efficiency, truncated RI-BPTT is better, while its accuracy is limited due to randomly initialized hidden states.
In terms of accuracy, on the other hand, the original truncated BPTT is better, while its memory usage is huge for training many long training videos.

For the good trade-off,
we must appropriately select which properties in the original BPTT and truncated RI-BPTT are essential for our goal.
The memory usage strongly depends on the video length (i.e., the number of frames) and the spatial dimension of the frame.
Since our goal is video-agnostic VSR that can work robustly for short and long videos, the hidden states of both the short and long videos are inevitably required.
On the other hand, the spatial dimension is reduced by cropping small windows, in general, for data augmentation.
To satisfy these requirements, our proposed method is designed as follows (Fig.~\ref{fig:PIBPTT}):

\noindent
{\bf Step 1:} At the beginning of each epoch, small windows are cropped in all $T$ frames of all $N$ training videos.
The cropped windows are located in the same image coordinates in all frames of each video, while their locations may differ between different videos.

\noindent
{\bf Step 2:} With a randomly initialized hidden state, all $T$ frames are fed into RNN for a one-time feed-forward process.
All computed hidden states (indicated by red boxes in Fig.~\ref{fig:PIBPTT}) are stored for Step 3.

\noindent
{\bf Step 3:} Assume that $l$ frames are recurrently processed for training RNN, as with truncated RNN.
In each video, given randomly selected $t$-th frame, the hidden state of $(t-1)$-th frame and $l$ frames from $t$-th frame to $(t+l-1)$-th frame are fed into unrolled RNN for training.
This training is repeated $R$ times in all $N$ videos in each epoch.
Note that $t$ may differ between different videos and different repeats.

These three steps are done in each epoch.
$R$ is a hyper-parameter.
As $R$ increases, truncated PI-BPTT becomes more efficient, while RNN overfits with the stored hidden states.
The effect of $R$ on computational efficiency and accuracy is validated in our experiments.

\subsection{Frame-number Conditioning for Difficulty-dependent VSR}
\label{subsection:frame_id}

Even with our PI-BPTT,
it is still difficult to super-resolve a long video.
A difficulty in VSR using RNN 
becomes higher as the number of recurrent steps increases.
To control VSR depending on the number of recurrent steps for improving the VSR quality, our method explicitly specifies the frame number of an input video by conditioning RNN with the frame number.

This conditioning is implemented as follows.
A one-channel image in which all pixel values are the frame number is concatenated with each input LR frame.
Since all pixel values in the LR frame are normalized between 0 and 1, those in the one-channel image are also normalized.
This normalized concatenated image is fed into RNN.

\section{Experimental Results}
\label{sec:experiments}

\subsection{Details}
\label{subsection:details}

\noindent
{\bf Datasets:}
The Vimeo dataset~\cite{DBLP:conf/cvpr/SajjadiVB18} and the training set in the REDS (REalistic and Dynamic Scenes) dataset~\cite{DBLP:conf/cvpr/NahBHMSTL19,DBLP:conf/cvpr/ChanZXL22a,DBLP:conf/cvpr/ChanWYDL21}, which is called REDStrain in what follows, are used as the training sets in our experiments (Table~\ref{table:train_data_statistics}).
However, in accordance with~\cite{DBLP:conf/cvpr/ChanZXL22a,DBLP:conf/cvpr/ChanWYDL21}, the REDS4 set consisting of the four videos of ID=000, 011, 015, and 020 is excluded from REDStrain and used as a test set.
In the Vimeo dataset, all videos are more static and longer than the REDS dataset.
In the REDS dataset, various scenes including a variety of objects, motions, and blurs are observed.

\begin{table}[t]
    \centering
    \caption{Training datasets.}
    \vspace*{-2mm}
    \scalebox{1.0}{
    \begin{tabular}{|l||c|c|c|} 
        \hline
        Datasets & \begin{tabular}{c}\# of\\ videos\end{tabular} & \begin{tabular}{c}\# of frames\\ per video\end{tabular} & Image size\\ 
        \hline \hline
        Vimeo & 278 & 120 & \begin{tabular}{c}360$\times$640 --\\ 1080$\times$2048\end{tabular} \\
        \hline
        REDStrain & 236 & 100 & 720$\times$1280 \\
        \hline
    \end{tabular}
    }
    \label{table:train_data_statistics}
    \vspace*{4mm}
    \caption{Validation datasets.}
    \vspace*{-2mm}
    \begin{tabular}{|l||c|c|c|} 
        \hline
        Datasets & \begin{tabular}{c}\# of\\ videos\end{tabular} & \begin{tabular}{c}\# of frames\\ per video\end{tabular} & Image size\\
        \hline \hline
        REDSval4 & 8 & 100 & 720$\times$1280 \\
        \hline
        Vimeoval4 & 8 & 74--100 & 540$\times$1024 \\
        \hline
    \end{tabular}
    \label{table:validation_data_statistics}
    \vspace*{4mm}
    \caption{Test datasets.}
    \vspace*{-2mm}
    \scalebox{1.0}{
    \begin{tabular}{|l||c|c|c|} 
        \hline
        Datasets & \begin{tabular}{c}\# of\\ videos\end{tabular} & \begin{tabular}{c} \# of frames\\per video \end{tabular} & Image size \\ 
        \hline \hline
        Vid4 & 4 & 34--49 & \begin{tabular}{c}480$\times$720--\\ 576$\times$720\end{tabular} \\
        \hline
        REDS4 & 4 & 100 & 720$\times$1280\\
        \hline
        Quasi-Static & 4 & 172--379 & \begin{tabular}{c}360$\times$640--\\ 540$\times$960\end{tabular}\\
        \hline
        Composit & 15 & 181--298 & \begin{tabular}{c}540$\times$960--\\ 720$\times$1280\end{tabular} \\
        \hline
    \end{tabular}
    }
    \label{table:test_data_statistics}
\end{table}

In accordance with~\cite{DBLP:conf/cvpr/ChanZXL22a,DBLP:conf/cvpr/ChanWYDL21}, four videos in the Vimeo datasets, which are called Vimeoval4, and REDSval4 are used as the validation sets for training the VSR models with the Vimeo and REDStrain datasets, respectively (Table~\ref{table:validation_data_statistics}).
The REDSval4 dataset consists of the four videos of ID=000, 001, 006, and 017 in the validation set of the REDS dataset.
In addition to these videos, static videos should be included in the validation set for training video-agnostic VSR.
As such static videos, the first frame of each of the above validation videos (i.e., four videos in each of REDSval4 and Vimeoval4) is temporally copied to synthesize a video.
In total, the four static validation videos are synthesized from each validation set.

To evaluate the video-agnostic performance of our method, a variety of videos, including various objects, static/dynamic, and short/long videos, should be included in the test set.
In the literature prior to~\cite{DBLP:conf/cvpr/ChicheWFS22}, however, only dynamic short videos (e.g., 32--100 frames including moving objects),
such as the Vid4\cite{DBLP:conf/cvpr/LiuS11} and REDS4\cite{DBLP:conf/cvpr/NahBHMSTL19}
datasets,
are used as test sets.
On the other hand, the Quasi-Static Video Set~\cite{DBLP:conf/cvpr/ChicheWFS22} consists of only static long videos (379 frames at maximum), which make it difficult to train RNN-based VSR, as validated in Sec.~\ref{section:preliminary}.

Therefore, our experiments use a combination of the aforementioned datasets as the test set (Table~\ref{table:test_data_statistics}).
As static long videos, the Quasi-Static Video Set is used. As short videos with various video properties, the Vid4~\cite{DBLP:conf/cvpr/LiuS11} and REDS4~\cite{DBLP:conf/cvpr/NahBHMSTL19}
datasets are used.
In addition to these videos,
long videos with various video properties are also required.
We synthesize such long videos by temporally concatenating videos included in the evaluation set of the REDS dataset and the SPMCS dataset~\cite{DBLP:conf/iccv/TaoGLWJ17}, each of whose original video lengths are short (e.g., between 31 and 100 frames).
This temporal video concatenation is done so that the original video and its reverse video are temporally concatenated as follows: $f_{1}, f_{2}, \cdots, f_{N-1}, f_{N}, f_{N-1}, \cdots, f_{2}, f_{1}, f_{2}, \cdots$, where $f_{i}$ denotes $i$-th frame of the original video.
With this concatenation process, the length of each synthesized video is between 181 and 298 frames.

\noindent
{\bf SR condition:}
Following experiments in the base VSR work~\cite{DBLP:conf/cvpr/SajjadiVB18,DBLP:conf/cvpr/ChanWYDL21}, the performance is evaluated for a factor of 4 VSR.
Each LR image is generated from its HR image by blurring it by the Gaussian ($\sigma=1.5$) and subsampling every four pixels, in accordance with~\cite{DBLP:conf/cvpr/ChicheWFS22,DBLP:conf/iccvw/FuoliGT19,DBLP:journals/tog/ChuXMLT20}.

\noindent
{\bf Training:}
In the training procedure of all experiments, the length of temporal random clipping is 15 frames, the size of spatial random cropping is $64\times64$, flipping and rotation processes are applied to the LR and HR images for data augmentation, the optimizer is Adam\cite{DBLP:journals/corr/KingmaB14}.

In experiments with FRVSR~\cite{DBLP:conf/cvpr/SajjadiVB18}, the mini-batch size is four, and the number of iterations is 500k.
The learning rate is $10^{-4}$, which is decreased by $\frac{1}{2}$ at 150k-th and 300k-th iterations.

In experiments with BasicVSR~\cite{DBLP:conf/cvpr/ChanWYDL21}, the mini-batch size is eight, and the number of iterations is 300k.
The optical flow estimator's and other networks' learning rates are $2.5\times10^{-4}$ and $10^{-4}$, respectively.
The learning rate is fixed during the first 5k iterations, while it is adjusted later by the Cosine Annealing~\cite{DBLP:conf/iclr/LoshchilovH17}.

\noindent
{\bf Evaluation metrics:}
The quality of the reconstructed video is measured as the mean of those of all frames in the video.
The frame quality is evaluated pixelwise by PSNR and SSIM.
PSNR and SSIM are measured on the Y channel of the YCbCr color space.

The computational efficiency is quantified by the computational time.
The training time is the mean time per iteration, including the time for image loading and all steps shown in Fig.~\ref{fig:RIBPTT} and Fig.~\ref{fig:PIBPTT}.
The training time in our method also includes the time for one-time feed-forward processing to compute a set of hidden states.
The feed-forward time added to the training time per iteration is $\frac{T_{ff}}{N_{it}}$, where $T_{ff}$ and $N_{it}$ denote the time for one-time feed-forward processing and the number of iterations in one epoch.
The training time is measured by Tesla V100-SXM2-32GB.

\subsection{Results}
\label{subsection:results}

\begin{table*}[t]
\centering
\caption{Quantitative results (PSNR/SSIM) of VSR. 
FRVSR and BasicVSR are trained by RI-BPTT (RI in the table), our proposed PI-BPTT (PI), and PI-BPTT and our proposed frame-number conditioning (FC).
The best and second-best PSNR results in each base method (i.e., FRVSR or BasicVSR) are colored red and blue.
}
\scalebox{1.0}{
\begin{tabular}{|l||c|c||c|c||c|c||c|c|}
\hline
                    & \multicolumn{4}{|c||}{Trained by Vimeo} & \multicolumn{4}{|c|}{Trained by REDStrain} \\ \cline{2-9}
                    & Vid4 & REDS4 & Quasi-Static & Composit & Vid4 & REDS4 & Quasi-Static & Composit \\ \hline\hline
MRVSR               & 26.93/0.828 & 29.62/0.830 & 32.26/0.880 & 29.45/0.821 & N/A & N/A & N/A & N/A \\ \hline\hline
FRVSR w/ RI         & 27.15/0.837 & \red{30.64}/0.861 & 30.43/0.854 & 29.94/0.837 & 27.11/0.841 & 31.93/0.889 & 24.81/0.680 & 29.19/0.816 \\ \hline
FRVSR w/ PI         & \blue{27.36}/0.846 & 30.63/0.862 & \blue{31.88}/0.874 & \blue{30.03}/0.842 & \blue{27.29}/0.848 & \red{31.99}/0.891 & \blue{25.23}/0.673 & \red{29.69}/0.829 \\ \hline
FRVSR w/ PI \& FC   & \red{27.39}/0.846 & \red{30.64}/0.862 & \red{31.94}/0.876 & \red{30.27}/0.845 & \red{27.37}/0.849 & \blue{31.98}/0.891 & \red{25.49}/0.618 & \blue{29.42}/0.816 \\ \hline\hline
BasicVSR w/ RI      & 27.95/0.868 & 32.59/0.906 & 23.23/0.679 & 29.07/0.831 & 28.14/0.871 & 33.57/0.921 & 20.90/0.438 & 28.86/0.770 \\ \hline
BasicVSR w/ PI      & \red{28.25}/0.873 & \blue{32.90}/0.911 & \blue{31.99}/0.873 & \red{31.76}/0.885 & \red{28.52}/0.878 & \blue{33.86}/0.925 & \blue{24.23}/0.638 & \red{31.03}/0.861 \\ \hline 
BasicVSR w/ PI \& FC& \blue{28.23}/0.873 & \red{32.91}/0.911 & \red{32.25}/0.878 & \blue{31.74}/0.886 & \blue{28.46}/0.878 & \red{33.87}/0.925 & \red{24.79}/0.605 & \blue{30.74}/0.847 \\ \hline
\end{tabular}
}
\label{table:quantitative_comparison}
\end{table*}

\begin{table}[t]
\centering
\caption{
The number of parameters (Mega), the training time per iteration (msec), and the inference time (msec).
}
\begin{tabular}{|l||c|c|c|}
\hline
& \# of params & T-Time & I-Time\\ \hline\hline
MRVSR               & 1.21 & N/A & 24 \\ \hline
FRVSR w/ RI         & \multirow{3}{*}{2.58} & 351 & \multirow{2}{*}{39} \\ \cline{1-1} \cline{3-3}
FRVSR w/ PI         &      & 1620 & \\ \cline{1-1} \cline{3-4}
FRVSR w/ PI \& FC   &      & 1621 & 37 \\ \hline\hline
BasicVSR w/ RI      & \multirow{3}{*}{6.29} & 881 & \multirow{2}{*}{94} \\ \cline{1-1} \cline{3-3}
BasicVSR w/ PI      &      & 2479 & \\ \cline{1-1} \cline{3-4} 
BasicVSR w/ PI \& FC&      & 2439 & 93 \\ \hline
\end{tabular}
\label{table:efficiency}
\end{table}

\subsubsection{Quantitative and Qualitative Results}

Quantitative results are shown in Table~\ref{table:quantitative_comparison}.
For comparison, the results of a SoTA VSR method (MRVSR~\cite{DBLP:conf/cvpr/ChicheWFS22}) and the base methods (FRVSR~\cite{DBLP:conf/cvpr/SajjadiVB18} and BasicVSR~\cite{DBLP:conf/cvpr/ChanWYDL21}) are also shown.
Since the authors' model of MRVSR is trained by the Vimeo dataset, no results are shown for the REDStrain set.
Note that the original FRVSR and BasicVSR are trained by RI-BPTT.
For the ablation study of our proposed method, FRVSR and BasicVSR are trained in two ways, namely trained only by PI-BPTT (``w/ PI'' in Table~\ref{table:quantitative_comparison}) and trained by PI-BPTT and frame-number conditioning (``w/ PI \& FC'' in Table~\ref{table:quantitative_comparison}).

\begin{figure}[t]
\begin{center}
\includegraphics[width=\columnwidth]{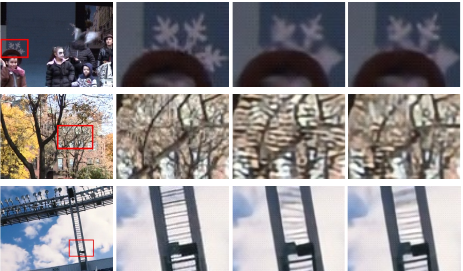}\\
GT\hspace*{15mm}
GT\hspace*{15mm}
w/ RI\hspace*{12mm}
w/ PI
\end{center}
\vspace*{-4mm}
\caption{Success cases in which PI-BPTT improves the VSR quality with FRVSR in short videos (i.e., Vid4 and REDS4). The window enclosed by each red rectangle in the leftmost image is zoomed-up. The ground truth (GT), the VSR image obtained with RI-BPTT (w/ RI), and the VSR image obtained with PI-BPTT (w/ PI) are zoomed-up.}
\label{fig:comparison_success}
\vspace*{-1mm}
\begin{center}
\includegraphics[width=\columnwidth]{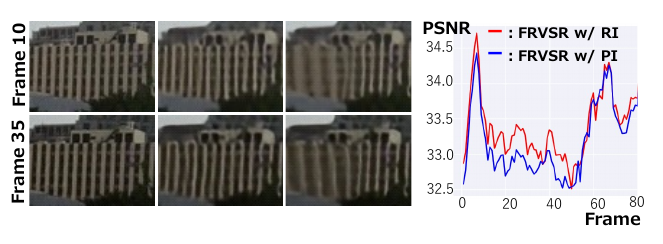}\\
\vspace*{-5mm}
GT\hspace*{8mm}
w/ RI\hspace*{8mm}
w/ PI\hspace*{27mm}~
\end{center}
\vspace*{-4mm}
\caption{Failure cases in which PI-BPTT degrades the VSR quality with FRVSR in short videos (i.e., Vid4 and REDS4).}
\label{fig:comparison_failure}
\end{figure}

\begin{figure}[t]
\begin{center}
\includegraphics[width=\columnwidth]{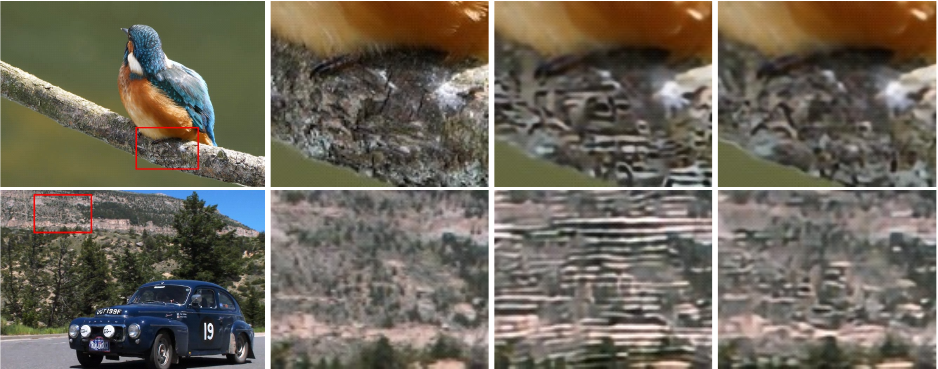}\\
~\vspace*{7mm}
GT\hspace*{17mm}
GT\hspace*{12mm}
w/ RI\hspace*{10mm}
w/ PI
\end{center}
\vspace*{-10mm}
\caption{FRVSR results for long videos (i.e., Quasi-static and Composit).}
\label{fig:comparison_success_long_FRVSR}
\vspace*{-2mm}
\end{figure}

In most cases in Table~\ref{table:quantitative_comparison}, our method improves the VSR quality.
This result demonstrates that our method effectively improves the VSR performance independently of video properties.
While only one exception is the case in which Vimeo and REDS4 are used as the training and test sets for FRVSR, respectively, all three results (i.e., ``FRVSR w/ RI, FRVSR w/ PI, and FRVSR w/ PI and FC) are almost equal.
While PI-BPTT improves the VSR quality in most cases (particularly for long video VSR with BasicVSR), our frame-number conditioning has less impact.

Our method also outperforms MRVSR by a large margin (e.g., 1.30, 3.29, and 2.29 PSNR gains by ``BasicVSR w/ PI \& FC'' in Vid4, REDS4, and Composit, respectively), while the PSNR scores of MRVSR and ``BasicVSR w/ PI \& FC'' are almost equal in the Quasi-Static set because MRVSR is optimized for static videos.

Examples in which our method improves the VSR quality for FRVSR are shown in Fig.~\ref{fig:comparison_success}.
On the other hand, Fig.~\ref{fig:comparison_failure} shows failure cases.
For a better understanding of the negative effect of our method, the temporal histories of PSNR are shown in the rightmost graph in Fig.~\ref{fig:comparison_failure}.
Comparison between the PSNR histories of FRVSR w/ RI-BPTT and w/ PI-BPTT reveals that the quality degradation caused at around the 10th frame is propagated to around the 35th frame in FRVSR w/ PI-BPTT.
Such degradation may be caused at an earlier frame because the effect of the random hidden state initialization in inference is trained better in RI-BPTT than in PI-BPTT.

\begin{figure}[t]
\begin{center}
\includegraphics[width=\columnwidth]{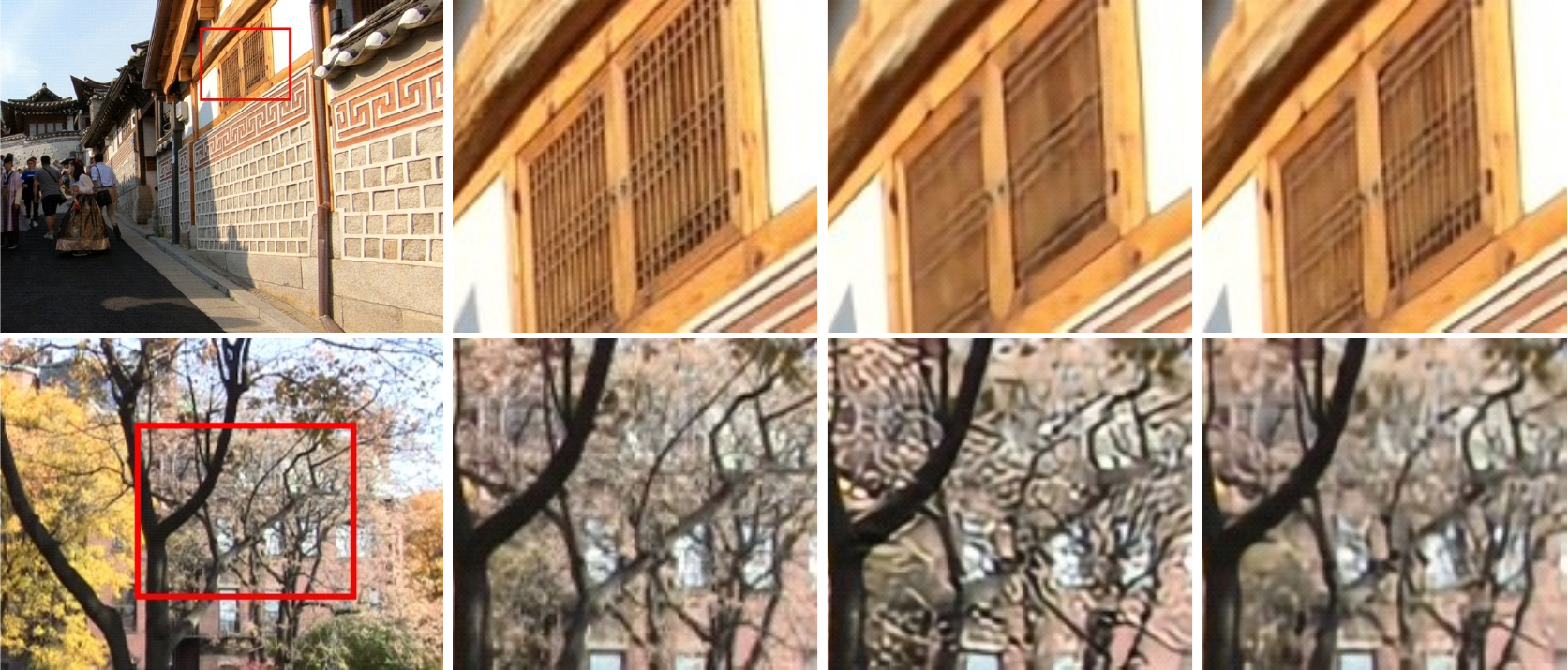}\\
~\vspace*{7mm}
GT\hspace*{17mm}
GT\hspace*{12mm}
w/ RI\hspace*{10mm}
w/ PI
\end{center}
\vspace*{-10mm}
\caption{Results of BasicVSR for short videos (i.e., Vid4 and REDS4).}
\label{fig:comparison_success_short_BasicVSR}
\vspace*{1mm}
\begin{center}
\includegraphics[width=\columnwidth]{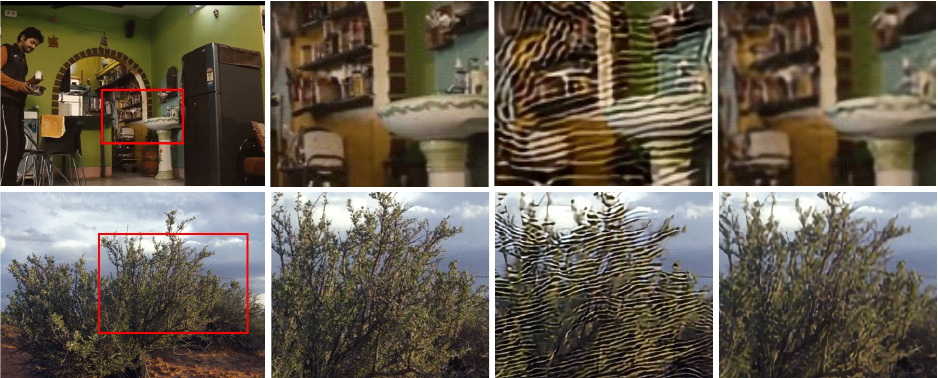}\\
~\vspace*{7mm}
GT\hspace*{17mm}
GT\hspace*{12mm}
w/ RI\hspace*{10mm}
w/ PI
\end{center}
\vspace*{-10mm}
\caption{Results of BasicVSR for long videos (i.e., Quasi-static and Composit).}
\label{fig:comparison_success_long_BasicVSR}
\end{figure}

The visual results reconstructed by BasicVSR for short and long videos are shown in Fig.~\ref{fig:comparison_success_short_BasicVSR} and Fig.~\ref{fig:comparison_success_long_BasicVSR}, respectively.
It can be seen that our method can improve the performance also with bidirectional RNN (i.e., BasicVSR).

\begin{figure}[t]
\begin{center}
\includegraphics[width=\columnwidth]{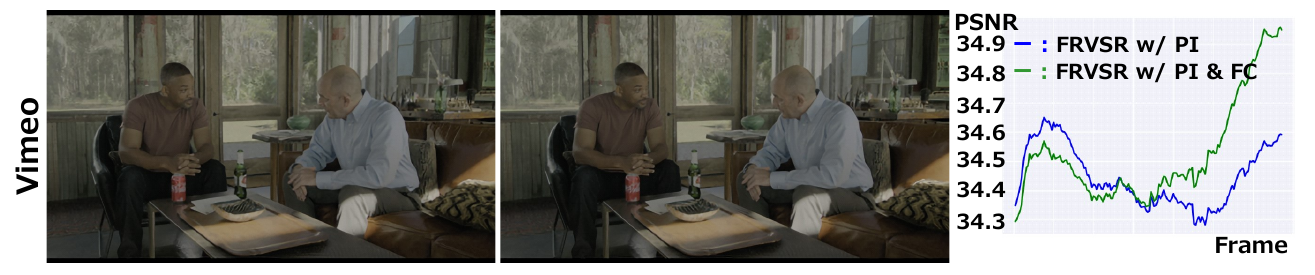}\\
\vspace*{-2mm}
{\scriptsize
w/ PI (PSNR=34.59)\hspace*{5mm}
w/ PI \& FC (PSNR=34.94)\hspace*{18mm}
}\\
\vspace*{2mm}
\includegraphics[width=\columnwidth]{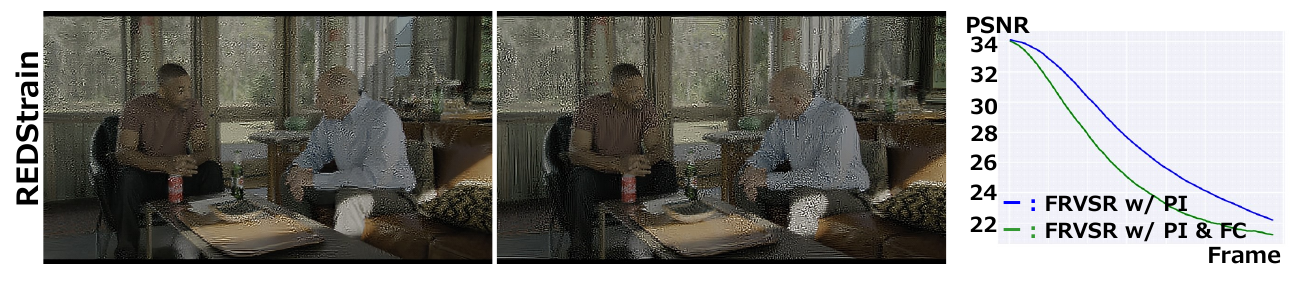}\\
\vspace*{-2mm}
{\scriptsize
w/ PI (PSNR=22.17)\hspace*{5mm}
w/ PI \& FC (PSNR=21.19)\hspace*{18mm}
}\\
\end{center}
\vspace*{-3mm}
\caption{Performance change depending on the training dataset. The top and bottom rows show the results obtained by the Vimeo and REDStrain training sets, respectively.}
\label{fig:comparison_train_FRVSR}
\end{figure}

As validated above, our method can improve the VSR quality.
However, the quality obtained by VSR trained by REDStrain is much worse than Vimeo on the Quasi-static video dataset.
For example, the PSNR scores obtained by Vimeo and REDStrain are 31.94 and 25.49, respectively, when ``FRVSR w/ PI \& FC'' is used.
The performance change depending on the training dataset is shown in Fig.~\ref{fig:comparison_train_FRVSR}.
With the Vimeo training set, which is shown at the top of Fig.~\ref{fig:comparison_train_FRVSR}, the quality is not decreased even as the frame number increases.
With the REDStrain training set, on the other hand, the quality gradually declines as the frame number increases.
This performance decline may be caused due to the large domain gap between the training and test sets.
While the REDStrain dataset includes many fast-motion and blurred images, no such images are observed in the Quasi-static video dataset.
This result suggests that a domain gap between the training test sets should be small, even if our method can learn various video properties in one RNN network.

\subsubsection{Trade-off between Efficiency and Accuracy}

\begin{table}[t]
\centering
\caption{Trade-off between efficiency (training time) and accuracy (PSNR) in FRVSR with changing $R$.
The best and second-best scores are colored red and blue on each
test
set.}
\scalebox{0.95}{
\begin{tabular}{|c||c|c|c|c|c|}
\hline
& Training time & Vid4 & REDS4 & Quasi-static & Composit \\ \hline\hline
Base & 351 & 27.15 & 30.64 & 30.43 & 29.94 \\ \hline\hline
$R=2$  & 1170 & \blue{27.37} & \red{30.68} & \blue{31.83} & \red{30.23}  \\ \hline
$R=4$  &  832 & 27.36 & \blue{30.66} & 31.38 & 30.09 \\ \hline
$R=8$  &  644 & 27.30 & \blue{30.66} & \red{31.91} & 30.16 \\ \hline
$R=16$ &  520 & \red{27.39} & 30.62 & 31.71 & 30.16  \\ \hline
$R=32$ &  413 & 27.32 & 30.63 & 31.49 & 30.10 \\ \hline
$R=48$ &  382 & 27.34 & 30.59 & 31.18 & 30.03 \\ \hline
$R=64$ &  363 & \blue{27.37} & 30.60 & 31.66 & \blue{30.22} \\ \hline
\end{tabular}
}
\label{table:tradeoff_FRVSR}
\end{table}

\begin{figure}[t]
  \begin{center}
     \includegraphics[width=\columnwidth]{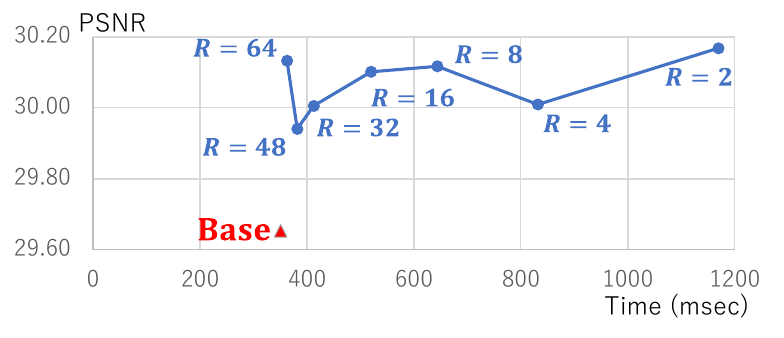}
     \vspace*{-6mm}
     \caption{Trade-off between efficiency and accuracy in FRVSR. The horizontal and vertical axes indicate the training time and PSNR, respectively.}
     \label{fig:tradeoff}
   \end{center}
\end{figure}

The effect of $R$ (i.e., the number of iterations in each epoch in PI-BPTT) on the trade-off between efficiency and accuracy is validated with FRVSR,
as shown in Table~\ref{table:tradeoff_FRVSR}.
For comparison, the results of the base method (i.e., FRVSR with RI-BPTT) are also shown.
Since the cost of computing the hidden states of RNN is reduced as $R$ increases, the training time is also reduced.
In terms of accuracy, it is expected that RNN overfits to the stored hidden states as $R$ increases because the same set of the hidden states are reused many times (i.e., $R$ times).
As expected, the best accuracy is obtained with the minimum $R$ (i.e., $R=2$) in the Vid4 and Composit datasets.
While the accuracy obtained with the minimum $R$ is the second best in the REDS4 and Quasi-Static datasets, the gap from the best score is insignificant.
Furthermore, the performance change depending on $R$ is insignificant in all the test sets.
For better understanding, the mean of all the four test sets are plotted in Fig.~\ref{fig:tradeoff}.
Indeed, the change in PSNR seems almost random within a small PSNR range (i.e., between 29.9 and 30.2).
On the other hand, the training time of $R=64$ can be reduced to almost that of the base method (i.e., FRVSR w/ RI-BPTT).
This result demonstrates that our method can improve the VSR quality with a slight increase in its training cost.

\section{Concluding Remarks}
\label{sec:conclusion}

This paper proposed a video-agnostic training scheme for RNN-based VSR.
While prior VSR methods train RNN by using short videos with randomly initialized hidden states,
such random hidden states in training differ from continuously updated hidden states in inference.
This domain gap
degrades the VSR quality.
To fill this domain gap, continuously updated hidden states are used for training RNN-based VSR in our proposed training scheme.
Since such continuously updated hidden states require huge memory and computational costs, a good trade-off between efficiency and accuracy is validated.

Important future work includes VSR quality evaluation in much longer videos.
It is impossible to completely avoid error accumulation in RNN, RNN should be reinitialized based on accumulated error for real applications~\cite{DBLP:journals/csur/LeeVL22}.

This work was partly supported by JSPS KAKENHI 22H03618.


\bibliographystyle{unsrt}
\bibliography{egbib}

\begin{thebibliography}{10}

\bibitem{DBLP:conf/iclr/MillerH19}
John Miller and Moritz Hardt.
\newblock Stable {R}ecurrent {M}odels.
\newblock In {\em ICLR}, 2019.

\bibitem{DBLP:conf/icml/MhammediHRB17}
Zakaria Mhammedi, Andrew~D. Hellicar, Ashfaqur Rahman, and James Bailey.
\newblock Efficient orthogonal parametrisation of recurrent neural networks using householder reflections.
\newblock In {\em ICML}, 2017.

\bibitem{DBLP:conf/icml/VorontsovTKP17}
Eugene Vorontsov, Chiheb Trabelsi, Samuel Kadoury, and Chris Pal.
\newblock On orthogonality and learning recurrent networks with long term dependencies.
\newblock In {\em ICML}, 2017.

\bibitem{DBLP:conf/icml/JoseCF18}
Cijo Jose, Moustapha Ciss{\'{e}}, and Fran{\c{c}}ois Fleuret.
\newblock Kronecker recurrent units.
\newblock In {\em ICML}, 2018.

\bibitem{DBLP:journals/pami/TanaySMDTLS23}
Thomas Tanay, Aivar Sootla, Matteo Maggioni, Puneet~K. Dokania, Philip H.~S. Torr, Ales Leonardis, and Gregory~G. Slabaugh.
\newblock Diagnosing and preventing instabilities in recurrent video processing.
\newblock {\em {IEEE} Trans. Pattern Anal. Mach. Intell.}, 45(2):1594--1605, 2023.

\bibitem{DBLP:conf/cvpr/ChicheWFS22}
Benjamin~Naoto Chiche, Arnaud Woiselle, Joana Frontera{-}Pons, and Jean{-}Luc Starck.
\newblock Stable {L}ong-{T}erm {R}ecurrent {V}ideo {S}uper-{R}esolution.
\newblock In {\em CVPR}, 2022.

\bibitem{DBLP:conf/cvpr/SajjadiVB18}
Mehdi S.~M. Sajjadi, Raviteja Vemulapalli, and Matthew Brown.
\newblock Frame-{R}ecurrent {V}ideo {S}uper-{R}esolution.
\newblock In {\em CVPR}, 2018.

\bibitem{DBLP:conf/cvpr/ChanWYDL21}
Kelvin C.~K. Chan, Xintao Wang, Ke~Yu, Chao Dong, and Chen~Change Loy.
\newblock Basic{VSR}: The {S}earch for {E}ssential {C}omponents in {V}ideo {S}uper-{R}esolution and {B}eyond.
\newblock In {\em CVPR}, 2021.

\bibitem{DBLP:conf/cvpr/ChanZXL22}
Kelvin C.~K. Chan, Shangchen Zhou, Xiangyu Xu, and Chen~Change Loy.
\newblock Investigating {T}radeoffs in {R}eal-{W}orld {V}ideo {S}uper-{R}esolution.
\newblock In {\em CVPR}, 2022.

\bibitem{DBLP:conf/cvpr/NahTGBHMSL19}
Seungjun Nah et~al.
\newblock {NTIRE} 2019 challenge on video super-resolution: Methods and results.
\newblock In {\em {CVPR} Workshops}, 2019.

\bibitem{DBLP:conf/eccv/FuoliHGTREKXLXW20}
Dario Fuoli et~al.
\newblock {AIM} 2020 challenge on video extreme super-resolution: Methods and results.
\newblock In {\em {ECCV} Workshops}, 2020.

\bibitem{DBLP:conf/cvpr/WangCYDL19}
Xintao Wang, Kelvin C.~K. Chan, Ke~Yu, Chao Dong, and Chen~Change Loy.
\newblock {EDVR:} {V}ideo {R}estoration {W}ith {E}nhanced {D}eformable {C}onvolutional {N}etworks.
\newblock In {\em CVPR}, 2019.

\bibitem{DBLP:conf/cvpr/TianZ0X20}
Yapeng Tian, Yulun Zhang, Yun Fu, and Chenliang Xu.
\newblock {TDAN:} {T}emporally-{D}eformable {A}lignment {N}etwork for {V}ideo {S}uper-{R}esolution.
\newblock In {\em CVPR}, 2020.

\bibitem{DBLP:journals/ijcv/XueCWWF19}
Tianfan Xue, Baian Chen, Jiajun Wu, Donglai Wei, and William~T. Freeman.
\newblock Video {E}nhancement with {T}ask-{O}riented {F}low.
\newblock {\em Int. J. Comput. Vis.}, 127(8):1106--1125, 2019.

\bibitem{DBLP:conf/cvpr/HarisSU19}
Muhammad Haris, Gregory Shakhnarovich, and Norimichi Ukita.
\newblock Recurrent {B}ack-{P}rojection {N}etwork for {V}ideo {S}uper-{R}esolution.
\newblock In {\em CVPR}, 2019.

\bibitem{DBLP:conf/cvpr/HarisSU20}
Muhammad Haris, Greg Shakhnarovich, and Norimichi Ukita.
\newblock Space-time-aware multi-resolution video enhancement.
\newblock In {\em CVPR}, 2020.

\bibitem{DBLP:conf/cvpr/ChanZXL22a}
Kelvin C.~K. Chan, Shangchen Zhou, Xiangyu Xu, and Chen~Change Loy.
\newblock Basic{VSR}++: Improving {V}ideo {S}uper-{R}esolution with {E}nhanced {P}ropagation and {A}lignment.
\newblock In {\em CVPR}, 2022.

\bibitem{DBLP:conf/iccv/YiWJJ019}
Peng Yi, Zhongyuan Wang, Kui Jiang, Junjun Jiang, and Jiayi Ma.
\newblock Progressive {F}usion {V}ideo {S}uper-{R}esolution {N}etwork via {E}xploiting {N}on-{L}ocal {S}patio-{T}emporal {C}orrelations.
\newblock In {\em ICCV}, 2019.

\bibitem{DBLP:conf/iccvw/FuoliGT19}
Dario Fuoli, Shuhang Gu, and Radu Timofte.
\newblock Efficient {V}ideo {S}uper-{R}esolution through {R}ecurrent {L}atent {S}pace {P}ropagation.
\newblock In {\em ICCV}, 2019.

\bibitem{DBLP:conf/eccv/IsobeJGLWT20}
Takashi Isobe, Xu~Jia, Shuhang Gu, Songjiang Li, Shengjin Wang, and Qi~Tian.
\newblock Video {S}uper-{R}esolution with {R}ecurrent {S}tructure-{D}etail {N}etwork.
\newblock In {\em ECCV}, 2020.

\bibitem{DBLP:journals/corr/abs-2201-12288}
Jingyun Liang, Jiezhang Cao, Yuchen Fan, Kai Zhang, Rakesh Ranjan, Yawei Li, Radu Timofte, and Luc~Van Gool.
\newblock {VRT:} {A} {V}ideo {R}estoration {T}ransformer.
\newblock {\em arXiv}, abs/2201.12288, 2022.

\bibitem{DBLP:journals/corr/abs-2106-06847}
Jiezhang Cao, Yawei Li, Kai Zhang, and Luc~Van Gool.
\newblock Video {S}uper-{R}esolution {T}ransformer.
\newblock {\em arXiv}, abs/2106.06847, 2021.

\bibitem{DBLP:conf/nips/VaswaniSPUJGKP17}
Ashish Vaswani, Noam Shazeer, Niki Parmar, Jakob Uszkoreit, Llion Jones, Aidan~N. Gomez, Lukasz Kaiser, and Illia Polosukhin.
\newblock Attention is {A}ll you {N}eed.
\newblock In {\em NIPS}, 2017.

\bibitem{DBLP:conf/nips/ShiGXWYD22}
Shuwei Shi, Jinjin Gu, Liangbin Xie, Xintao Wang, Yujiu Yang, and Chao Dong.
\newblock Rethinking alignment in video super-resolution transformers.
\newblock In {\em NeurIPS}, 2022.

\bibitem{DBLP:conf/nips/LiangFXRIGC0TG22}
Jingyun Liang, Yuchen Fan, Xiaoyu Xiang, Rakesh Ranjan, Eddy Ilg, Simon Green, Jiezhang Cao, Kai Zhang, Radu Timofte, and Luc~Van Gool.
\newblock Recurrent video restoration transformer with guided deformable attention.
\newblock In {\em NeurIPS}, 2022.

\bibitem{DBLP:conf/eccv/ZhangFL22}
Kaidong Zhang, Jingjing Fu, and Dong Liu.
\newblock Flow-guided transformer for video inpainting.
\newblock In {\em ECCV}, 2022.

\bibitem{DBLP:conf/icml/LinCHWYZDZTG22}
Jing Lin, Yuanhao Cai, Xiaowan Hu, Haoqian Wang, Youliang Yan, Xueyi Zou, Henghui Ding, Yulun Zhang, Radu Timofte, and Luc~Van Gool.
\newblock Flow-guided sparse transformer for video deblurring.
\newblock In {\em ICML}, 2022.

\bibitem{DBLP:conf/iclr/SedghiGL19}
Hanie Sedghi, Vineet Gupta, and Philip~M. Long.
\newblock The singular values of convolutional layers.
\newblock In {\em ICLR}, 2019.

\bibitem{DBLP:conf/iclr/MiyatoKKY18}
Takeru Miyato, Toshiki Kataoka, Masanori Koyama, and Yuichi Yoshida.
\newblock Spectral {N}ormalization for {G}enerative {A}dversarial {N}etworks.
\newblock In {\em ICLR}, 2018.

\bibitem{DBLP:conf/iclr/SanyalTD20}
Amartya Sanyal, Philip H.~S. Torr, and Puneet~K. Dokania.
\newblock Stable {R}ank {N}ormalization for {I}mproved {G}eneralization in {N}eural {N}etworks and {G}ans.
\newblock In {\em ICLR}, 2020.

\bibitem{DBLP:journals/ml/GoukFPC21}
Henry Gouk, Eibe Frank, Bernhard Pfahringer, and Michael~J. Cree.
\newblock Regularisation of neural networks by enforcing lipschitz continuity.
\newblock {\em Mach. Learn.}, 110(2):393--416, 2021.

\bibitem{DBLP:journals/pieee/Werbos90}
Paul~J. Werbos.
\newblock Backpropagation through time: what it does and how to do it.
\newblock {\em Proc. {IEEE}}, 78(10):1550--1560, 1990.

\bibitem{DBLP:conf/uai/AicherFF19}
Christopher Aicher, Nicholas~J. Foti, and Emily~B. Fox.
\newblock Adaptively truncating backpropagation through time to control gradient bias.
\newblock In {\em UAI}, 2019.

\bibitem{DBLP:conf/slt/TangG18}
Hao Tang and James~R. Glass.
\newblock On training recurrent networks with truncated backpropagation through time in speech recognition.
\newblock In {\em 2018 {IEEE} Spoken Language Technology Workshop, {SLT}}, 2018.

\bibitem{DBLP:journals/corr/TallecO17a}
Corentin Tallec and Yann Ollivier.
\newblock Unbiasing truncated backpropagation through time.
\newblock {\em arXiv}, abs/1705.08209, 2017.

\bibitem{DBLP:journals/neco/WilliamsP90}
Ronald~J. Williams and Jing Peng.
\newblock An efficient gradient-based algorithm for on-line training of recurrent network trajectories.
\newblock {\em Neural Comput.}, 2(4):490--501, 1990.

\bibitem{DBLP:conf/cvpr/NahBHMSTL19}
Seungjun Nah et~al.
\newblock {NTIRE} 2019 {C}hallenge on {V}ideo {D}eblurring and {S}uper-{R}esolution: {D}ataset and {S}tudy.
\newblock In {\em {CVPR} Workshops}, 2019.

\bibitem{DBLP:conf/cvpr/LiuS11}
Ce~Liu and Deqing Sun.
\newblock A {B}ayesian approach to adaptive video super resolution.
\newblock In {\em CVPR}, 2011.

\bibitem{DBLP:conf/iccv/TaoGLWJ17}
Xin Tao, Hongyun Gao, Renjie Liao, Jue Wang, and Jiaya Jia.
\newblock Detail-{R}evealing {D}eep {V}ideo {S}uper-{R}esolution.
\newblock In {\em ICCV}, 2017.

\bibitem{DBLP:journals/tog/ChuXMLT20}
Mengyu Chu, You Xie, Jonas Mayer, Laura Leal{-}Taix{\'{e}}, and Nils Thuerey.
\newblock Learning temporal coherence via self-supervision for {G}an-based video generation.
\newblock {\em {ACM} Trans. Graph.}, 39(4):75, 2020.

\bibitem{DBLP:journals/corr/KingmaB14}
Diederik~P. Kingma and Jimmy Ba.
\newblock Adam: {A} {M}ethod for {S}tochastic {O}ptimization.
\newblock In {\em ICLR}, 2015.

\bibitem{DBLP:conf/iclr/LoshchilovH17}
Ilya Loshchilov and Frank Hutter.
\newblock {SGDR:} {S}tochastic {G}radient {D}escent with {W}arm {R}estarts.
\newblock In {\em ICLR}, 2017.

\bibitem{DBLP:journals/csur/LeeVL22}
Royson Lee, Stylianos~I. Venieris, and Nicholas~D. Lane.
\newblock Deep neural network-based enhancement for image and video streaming systems: {A} survey and future directions.
\newblock {\em {ACM} Comput. Surv.}, 54(8):169:1--169:30, 2022.

\end{thebibliography}

\end{document}